\documentclass[final,5p,times,twocolumn]{elsarticle}

\usepackage{multirow}
\usepackage{threeparttable}
\usepackage{float}
\usepackage{amssymb}
\usepackage{graphicx}
\usepackage{amsmath}
\usepackage{url}
\usepackage{lineno,hyperref}
\usepackage{url}
\modulolinenumbers[5]

\journal{Engineering applications of artificial intelligence}

\begin{document}
\begin{frontmatter}

% \mainmatter  % start of an individual contribution

% first the title is needed
\title{Cross-Modality Synthesis from CT to PET \\
using FCN and GAN Networks \\
for Improved Automated Lesion Detection }

\author[tau]{Avi Ben-Cohen\corref{cor1}}
\ead{avibenc@mail.tau.ac.il}
\author[sheba]{Eyal Klang}
\author[sheba]{Stephen P. Raskin}
\author[sheba]{Shelly Soffer}
\author[sheba2,UCL]{Simona Ben-Haim}
\author[sheba]{Eli Konen}
\author[sheba]{Michal Marianne Amitai\fnref{shared}}
\author[tau]{Hayit Greenspan\fnref{shared}}
\fntext[shared]{These authors contributed equally.}
\cortext[cor1]{Corresponding author.}

\address[tau]{Tel Aviv University, Faculty of Engineering, Department of Biomedical Engineering, Medical Image Processing Laboratory, Tel Aviv 69978, Israel}
\address[sheba]{Sheba Medical Center, Diagnostic Imaging Department, Abdominal Imaging Unit, affiliated to Sackler school of medicine Tel Aviv University, Tel Hashomer 52621, Israel}
\address[sheba2] {Sheba Medical Center, Department of Nuclear Medicine, Tel Hashomer 52621, Israel}
\address[UCL] {Institute of Nuclear Medicine, University College London Hospital, London NW1 2BU, UK}

\begin{abstract}
In this work we present a novel system for generation of virtual PET images using CT scans. We combine a fully convolutional network (FCN) with a conditional generative adversarial network (GAN) to generate simulated PET data from given input CT data.
The synthesized PET can be used for false-positive reduction in lesion detection solutions. Clinically, such solutions may enable lesion detection and drug treatment evaluation in a CT-only environment, thus reducing the need for  the more expensive and radioactive PET/CT scan.   Our dataset includes 60 PET/CT scans from Sheba Medical center.  We used 23 scans for training and 37 for testing. Different schemes to achieve the synthesized output were qualitatively compared. Quantitative evaluation was conducted using an  existing lesion detection software, combining the synthesized PET as a false positive reduction layer for the detection of malignant lesions in the liver. Current results look promising showing a 28\% reduction in the average false positive per case from 2.9 to 2.1.   The suggested solution is comprehensive and can be expanded to additional body organs, and different modalities. 
\end{abstract}


\end{frontmatter}
% \linenumbers

\section{Introduction}
\begin{figure}[H]
\centering
\includegraphics[height=3.0 cm]{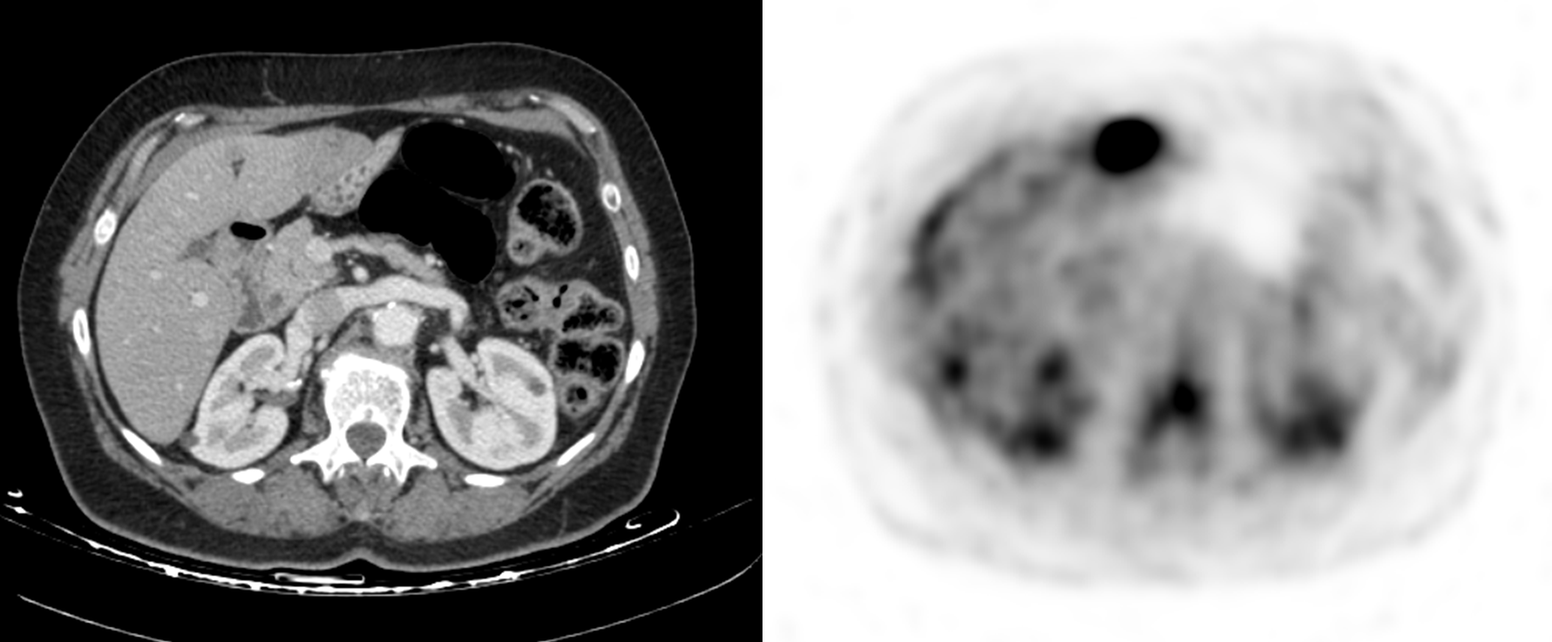}
\caption{An axial CT slice (left) with its corresponding PET slice (right). Dark regions in the PET image indicate high FDG uptake.}
\label{fig:pair}
\end{figure}

The combination of positron emission tomography (PET) and computerized tomography (CT) scanners has become a standard component of diagnosis and staging in oncology \cite{Weber,Kelloff}. An increased accumulation of Fluoro-D-glucose (FDG) in PET relative to normal tissue is a useful marker for many cancers and can help in detection and localization of malignant lesions \cite{Kelloff}. Additionally, PET/CT imaging is becoming an important evaluation tool for new drug therapies \cite{Weber2}. An example of an axial slice taken from a CT scan and its corresponding PET slice is shown in Figure \ref{fig:pair}. It can be seen that the PET image resolution is lower than the CT image resolution showing less anatomical details. However, there is a malignant liver lesion that is less visible in the CT image and can be easily detected in the PET image as a large dark blob.

Although PET imaging has many advantages and its use is steadily increasing, it has a few disadvantages. PET/CT entails added radiation exposure in comparison to CT-only scans. Moreover, PET/CT is relatively expensive compared to CT. Hence, it is still not offered in the large proportion of medical centers in the world. 
The clinical importance of PET in the management of cancer patients and on the other hand the difficulty in providing PET imaging as part of standard imaging,  raises a potential need for an alternative, less expensive, fast, and easy to use PET-like imaging.

Several works had recently explored cross-modality synthesis using deep learning methods \cite{Nie,Han,Xiang}. In these works, different deep learning based methods and architectures were explored to learn an end-to-end nonlinear mapping from magnetic resonance images to CT images. For the case of unpaired data, a CycleGAN model was used to synthesize brain CT images from brain MR by Wolterink et al. \cite{Wolterink}. Chartsias et al. \cite{Chartsias} demonstrated a similar concept for synthesizing cardiac MR images from CT images. In the case of PET/CT pairs, the PET study can be used to highlight malignant lesions and improve the detection compared to the use of CT data alone. Bi et al. \cite{Bi} used a multi-channel generative adversarial network that synthesizes PET images from CT images with manually annotated lung tumors. Their model learns the integration from both CT
and a given annotated label, to synthesize the high uptake and the anatomical background. They have demonstrated using the synthesized PET images a comparable detection performance to that achieved using the original PET data. We note that manual labeling of the tumors is needed in this work.

In the current work  our objective is to use information from CT data to estimate PET-like images with an emphasis on malignant lesions in the liver. The suggested system is fully automated, with no manual labeling needed. Similar to the radiologists, who have an easier time identifying malignant liver lesions in a PET/CT scan (vs only a CT scan) 
we want to make use of the estimated PET-like images to improve the detection of malignant lesions using an automated lesion detection software. 

The proposed system is based on a fully convolutional network (FCN) and a conditional GAN (cGAN). The contributions of this work include: 1) We present a novel method to synthesize PET images from CT images, focused on malignant lesions with no manually labeled data; 2) The synthesized PET is shown to improve an existing automatic lesion detection software; 3) Reconstruction measures are presented for comparison between different methods.

This work is an extension to earlier work  \cite{PET}, in which we used a pyramid based image blending step  to combine the advantages of an FCN and a cGAN network.
In the current work we present a novel system architecture that obviates the need for an image blending step, thus providing savings in time and reducing the need for manually defining a threshold for the blending mask, while improving the system performance (as will be demonstrated in section \ref{sec:experiments}). In addition, the dataset was substantially extended. 

To achieve the virtual PET we use advanced deep learning techniques with both fully convolutional networks and conditional adversarial networks as described in the following subsections.

\subsection{Fully Convolutional Networks}
In recent years, deep learning has become a dominant research topic in numerous fields. Specifically, Convolutional Neural Networks (CNN) have been used for many challenges in computer vision. CNN obtained outstanding performance on different tasks, such as visual object recognition, image classification, hand-written character recognition and more. Deep CNNs introduced by LeCun et al. \cite{LeCun}, is a supervised learning model formed by multi-layer neural networks.
CNNs are fully data-driven and can retrieve hierarchical features automatically by building high-level features from low-level ones, thus obviating the need to manually customize hand-crafted features. Previous works have shown the benefit of using a fully convolutional architecture for liver lesion detection and segmentation applications \cite{FCN,Christ}. FCNs can take input of arbitrary size and produce correspondingly-sized output with efficient inference and learning. Unlike patch based methods, the loss function using this architecture is computed over the entire image. The network processes entire images instead of patches, which removes the need to select representative patches, eliminates redundant calculations where patches overlap, and therefore scales up more efficiently with image resolution. Moreover, there is a fusion of different scales by adding links that combine the final prediction layer with lower layers with finer strides.

\subsection{Conditional Adversarial Networks}
\begin{figure}
\centering
\includegraphics[width=0.5\textwidth]{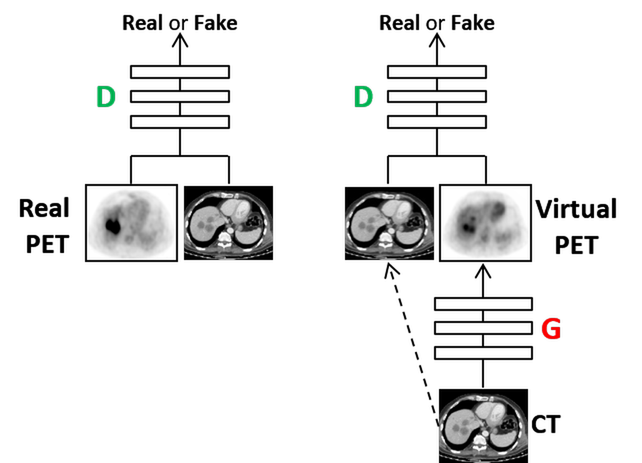}

\caption{Training a cGAN to predict PET images from
CT images. The discriminator, D, learns to classify between real and
synthesized pairs. The generator, G, learns to fool the discriminator.}
\label{fig:gan}
\end{figure}

More recent works show the use of Generative Adversarial Networks (GANs) for image to image translation \cite{Isola}. GANs are generative models that learn a mapping from random noise vector z to output image y \cite{Goodfellow}. In contrast, conditional GANs (cGANs) learn a mapping from observed image x and random noise vector z, to y. The generator G is trained to produce outputs that cannot be distinguished from “real” images by an adversarially trained discriminator, D, which is trained to detect real vs fake images. 
Figure \ref{fig:gan} shows a diagram of this procedure.
\\

In this study we use FCN and cGAN to generate PET-like images from CT volumes. The strengths of both methods are used to create realistic looking virtual PET images. We focus our attention to hepatic malignant lesions. The method is presented in Section \ref{sec:Methods}.  Experiments and results are described in Section \ref{sec:experiments}. The experiments include comparison of various algorithmic solutions to the task along with an evaluation of the benefit of our method to an existing automatic liver lesion detection software. We conclude this paper with a discussion in Section \ref{sec:discussion}.

\section{Methods}
\label{sec:Methods}
\begin{figure}
\centering
\includegraphics[width=0.5\textwidth]{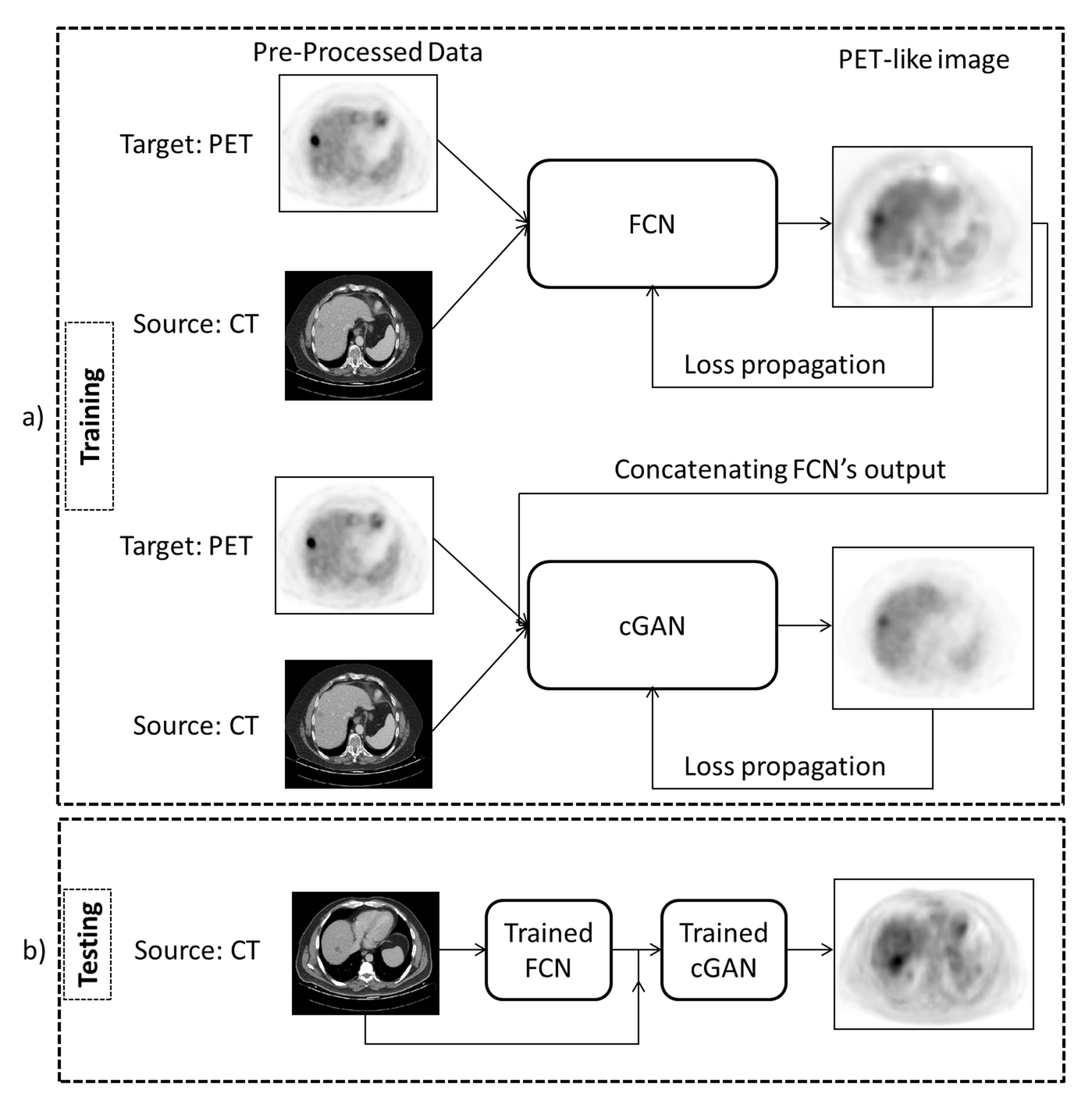}
\caption{The proposed virtual PET system.}
\label{fig:diagram}
\end{figure}

Our framework includes two main modules: a training module which includes data preparation, and a testing module which accepts CT images as input and predicts synthesized PET images. We use an FCN to generate initial PET-like images given the input CT images. We next use a cGAN to improve and refine the FCN output. Figure \ref{fig:diagram} shows a diagram of our general framework. Each module will be described in depth in the following subsections.

\subsection{Training Data Preparation}
The training input for the FCN includes two image types: a source CT image  and a target PET image. A similar size is needed for the two images and in most cases the PET resolution is much lower than the CT.
Hence, the first step in preparing the data for training was to align the PET scans with the CT scans using the given offset (provided for each scan) and the voxel size (in $mm$) ratio between both scans. Let us denote $T = (t_x, t_y, t_z)$ as the given offset between the CT and the PET scan, $S_{ct} = (h_{ct},w_{ct}, d_{ct})$ as the CT voxel size, and $S_{pet} = (h_{pet},w_{pet}, d_{pet})$ as the PET voxel size with $h$, $w$, $d$ representing the size in each dimension. Note that larger voxel size means lower resolution. Next, the following affine transformation with linear interpolation is used to align the PET scan to the CT scan:
\[
A
=
\begin{bmatrix}
    h_{pet}/h_{ct} & 0             & 0               & t_x \\
    0              & w_{pet}/w_{ct} & 0              & t_y \\
    0              & 0              & d_{pet}/d_{ct} & t_z \\
    0              & 0         		& 0  			 & 1
\end{bmatrix}
\]

The standardized uptake value (SUV) is commonly used as a relative measure of FDG uptake \cite{Higashi} as in equation \ref{eq:suv}:
\begin{equation} 
\label{eq:suv}
SUV=\frac{r}{a'/w}
\end{equation}
where $r$ is the radioactivity concentration [kBq/ml] measured by the PET scanner within a region of interest (ROI), $a'$ is the decay-corrected amount of injected radiolabeled FDG [kBq], and $w$ is the weight of the patient [g], which is used as a surrogate for a distribution volume of tracer.

CT and PET studies include a large value range. To assist the network to  learn the translation between these modalities, we found experimentally that some constraints were helpful: 
we used contrast adjustment, by clipping extreme values and scaling, to adjust the PET images into the SUV range of 0 to 20. This range includes most of the interesting SUV values of malignant lesions. Similarly, CT image values were adjusted to be within -160 HU to 240 HU (Hounsfield Units);  the standard HU windowing used by the radiologists when evaluating the liver parenchyma. Let us denote the minimum value of interest in a given scan $I$ as $I_{min}$ and the maximum value of interest as $I_{max}$. These extreme values are clipped as follows:
\begin{equation}
    \tilde{I}=
    \begin{cases}
      I_{min}, & \text{if}\ I<I_{min} \\
      I_{max}, & \text{if}\ I>I_{max}\\ 
      I, & \text{otherwise}
    \end{cases}
\end{equation}
where $\tilde{I}$ is the result of the clipping operation.
Additionally, each scan values were linearly adjusted to $[0,1]$ as in:
\begin{equation}
I_{final} = (\tilde{I}-I_{min})/(I_{max}-I_{min})
\end{equation}

\subsection{Fully Convolutional Network Architecture}
\begin{figure}
\centering
\includegraphics[width=0.5\textwidth]{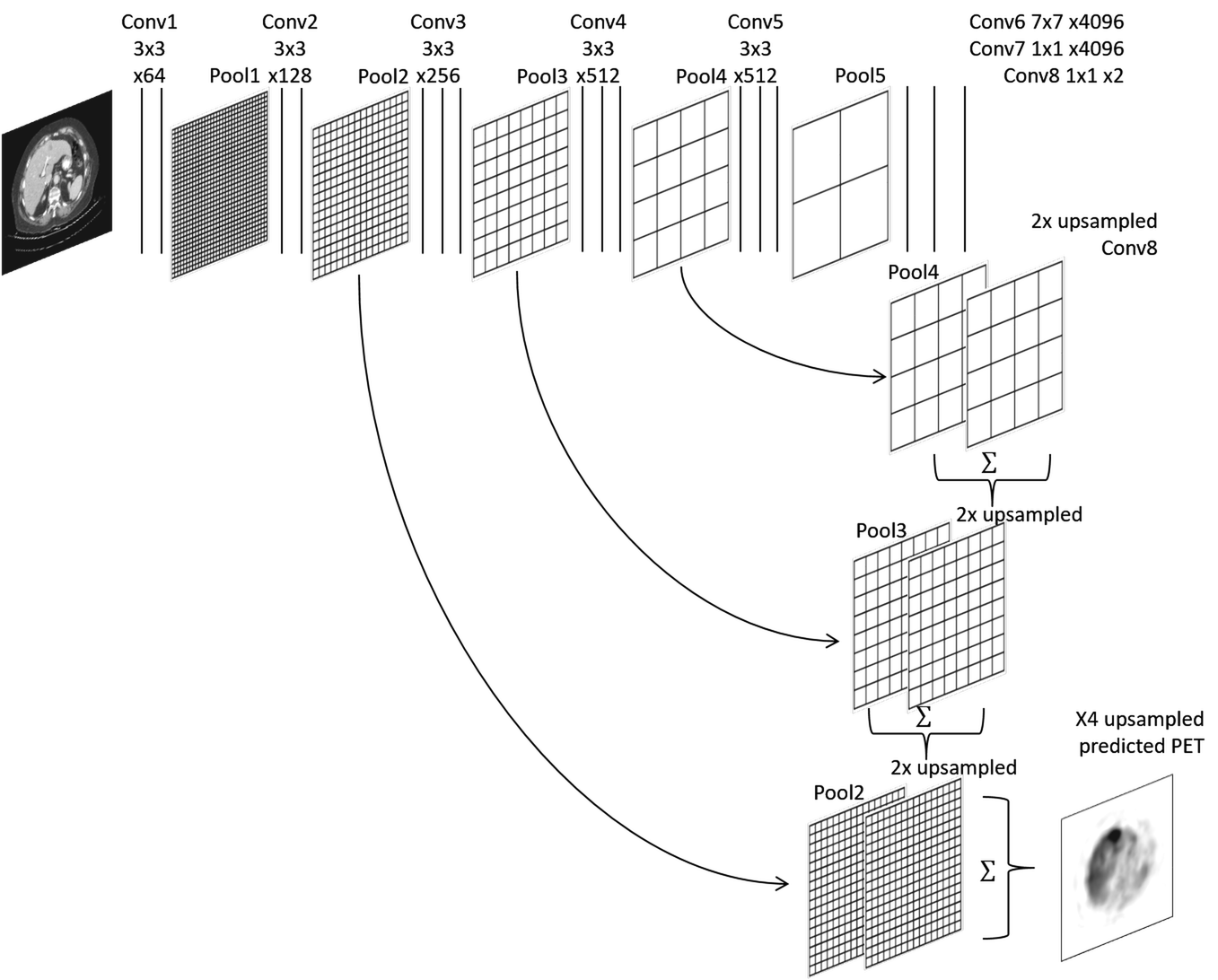}
\caption{FCN-4s architecture. Each convolution layer is illustrated by a straight line with the receptive field size and number of channels denoted above. The ReLU activation function and drop-out are not shown for brevity.}
\label{fig:FCN}
\end{figure}

In the following we describe the FCN used for both training and testing as in Figure \ref{fig:diagram}a and \ref{fig:diagram}b. The FCN network architecture uses the VGG 16- layer net \cite{Simonyan}. We convert all fully connected layers to convolutions and remove the classification layer. We append a 1x1 convolution with channel dimension to generate the PET-like images. Upsampling is performed in-network for end-to-end learning by backpropagation from the pixelwise $L_2$ loss. The FCN-4s net was used as our network, which learned to combine coarse, high layer information with fine, low layer information as described in \cite{Shelhamer} with an additional skip connection by linking the Pool2 layer in a similar way to the linking of the Pool3 and Pool4 layers in Figure \ref{fig:FCN}.

\subsection{cGAN Architecture}
The output from the FCN was  found experimentally to have a good response in regions with high SUV but to be less accurate and blurry in  regions with low contrast. 
Hence, cGAN was used to refine the FCN's output. The input to the cGAN included two channels, one with the CT image and the second with the corresponding FCN output (simple concatenation). The training concept follows the same trend of the training process of the FCN optimizing the network's generator based on the pixelwise $L_2$ loss as well as the cross entropy classification error for the discriminator, as will be further elaborated in section \ref{sec:Loss}. We adapt a similar cGAN architecture as in \cite{Isola} with a few modifications. In the original cGAN the objective can be expressed as:

\begin{align}
\label{eq:cGAN_orig}
\mathcal{L}_{cGAN}(G,D) = &\mathbb{E}_{ct,pet}[\log D(ct,pet)] + \nonumber \\ &\mathbb{E}_{ct,z}[\log (1-D(ct,G(ct,z))]
\end{align}
where $G$ tries to minimize this objective against an adversarial $D$ that tries to maximize it, $ct$ is the CT input slice, $pet$ is the corresponding PET slice, and $z$ is a Gaussian random noise variable, in the range of  [-0.005, 0.005], that was added to the CT input slice in each epoch. In our objective we embed the FCN's output ($fcn$) as in:
\begin{align}
\label{eq:cGAN_modified}
\mathcal{L}_{Modified-cGAN}(G,D) = &\mathbb{E}_{fcn,ct,pet}[\log D(fcn,ct,pet)] + \nonumber \\ &\mathbb{E}_{fcn,ct,z}[\log (1-D(fcn,ct,G(fcn,ct,z))]
\end{align}

We tried both $L1$ and $L2$ distance measures for the generator $G$. No noticeable difference was observed using the different distance measures and we chose to use the $L2$ in our experiments. The final optimization process:

\begin{align}
\label{eq:final_optimization}
G^*=&{arg \min }_{G} {\max }_{D} L_{Modified-cGAN}(G,D)+ \nonumber \\ &\lambda \mathbb{E}_{fcn,ct,z,pet}\Vert pet-G(fcn,ct,z)\Vert _2
\end{align}

where $G^*$ is the optimal setting and $\lambda$ balances the contribution of the two terms.

Table \ref{tab:generator} and Table \ref{tab:discriminator} show the different components of the network for the "U-Net" \cite{Ronneberger} based generator (encoder and decoder) and the discriminator respectively.

\begin{table}
\centering
\caption{The "U-Net" based generator used in the proposed cGAN architecture.}
\label{tab:generator}
\resizebox{\columnwidth}{!}{%
\begin{tabular}{|c|c|c|c|c|c|}
\hline
\multicolumn{3}{|c|}{\textbf{U-Net encoder}}                                                                                                           & \multicolumn{3}{c|}{\textbf{U-Net decoder}}                                                                                                                            \\ \hline
\textbf{Layer} & \textbf{Details}                                                             & \textbf{Size}                                          & \textbf{Layer} & \textbf{Details}                                                                             & \textbf{Size}                                          \\ \hline
input          &    \begin{tabular}[c]{@{}c@{}}CT image;\\ FCN's output\end{tabular}                                                                          & \begin{tabular}[c]{@{}c@{}}512x512\\ x2\end{tabular}   & upsampling1    & \begin{tabular}[c]{@{}c@{}}2x2 upsample of conv5\_2\\ concatenate with conv4\_2\end{tabular} & \begin{tabular}[c]{@{}c@{}}64x64\\ x768\end{tabular}   \\ \hline
conv1\_1       & \begin{tabular}[c]{@{}c@{}}3x3x32; dilaton rate 3; \\ LeakyReLU\end{tabular} & \begin{tabular}[c]{@{}c@{}}512x512\\ x32\end{tabular}  & conv6\_1       & \begin{tabular}[c]{@{}c@{}}3x3x256;\\  LeakyReLU\end{tabular}                                & \begin{tabular}[c]{@{}c@{}}64x64\\ x256\end{tabular}   \\ \hline
conv1\_2       & \begin{tabular}[c]{@{}c@{}}3x3x32; dilaton rate 3; \\ LeakyReLU\end{tabular} & \begin{tabular}[c]{@{}c@{}}512x512\\ x32\end{tabular}  & conv6\_2       & \begin{tabular}[c]{@{}c@{}}3x3x256; \\ LeakyReLU\end{tabular}                                & \begin{tabular}[c]{@{}c@{}}64x64\\ x256\end{tabular}   \\ \hline
pool1          & 2x2 max pool; stride 2                                                       & \begin{tabular}[c]{@{}c@{}}256x256\\ x32\end{tabular}  & upsampling2    & \begin{tabular}[c]{@{}c@{}}2x2 upsample of conv6\_2\\ concatenate with conv3\_2\end{tabular} & \begin{tabular}[c]{@{}c@{}}128x128\\ x384\end{tabular} \\ \hline
conv2\_1       & \begin{tabular}[c]{@{}c@{}}3x3x64; dilaton rate 2; \\ LeakyReLU\end{tabular} & \begin{tabular}[c]{@{}c@{}}256x256\\ x64\end{tabular}  & conv7\_1       & \begin{tabular}[c]{@{}c@{}}3x3x128;\\ LeakyReLU\end{tabular}                                 & \begin{tabular}[c]{@{}c@{}}128x128\\ x128\end{tabular} \\ \hline
conv2\_2       & \begin{tabular}[c]{@{}c@{}}3x3x64; dilaton rate 2;\\  LeakyReLU\end{tabular} & \begin{tabular}[c]{@{}c@{}}256x256\\ x64\end{tabular}  & conv7\_2       & \begin{tabular}[c]{@{}c@{}}3x3x128; \\ LeakyReLU\end{tabular}                                & \begin{tabular}[c]{@{}c@{}}128x128\\ x128\end{tabular} \\ \hline
pool2          & 2x2 max pool; stride 2                                                       & \begin{tabular}[c]{@{}c@{}}128x128\\ x64\end{tabular}  & upsampling3    & \begin{tabular}[c]{@{}c@{}}2x2 upsample of conv7\_2\\ concatenate with conv2\_2\end{tabular} & \begin{tabular}[c]{@{}c@{}}256x256\\ x192\end{tabular} \\ \hline
conv3\_1       & \begin{tabular}[c]{@{}c@{}}3x3x128; \\ LeakyReLU\end{tabular}                & \begin{tabular}[c]{@{}c@{}}128x128\\ x128\end{tabular} & conv8\_1       & \begin{tabular}[c]{@{}c@{}}3x3x64; dilaton rate 2;\\  LeakyReLU\end{tabular}                 & \begin{tabular}[c]{@{}c@{}}256x256\\ x64\end{tabular}  \\ \hline
conv3\_2       & \begin{tabular}[c]{@{}c@{}}3x3x128; \\ LeakyReLU\end{tabular}                & \begin{tabular}[c]{@{}c@{}}128x128\\ x128\end{tabular} & conv8\_2       & \begin{tabular}[c]{@{}c@{}}3x3x64; dilaton rate 2; \\ LeakyReLU\end{tabular}                 & \begin{tabular}[c]{@{}c@{}}256x256\\ x64\end{tabular}  \\ \hline
pool3          & 2x2 max pool; stride 2                                                       & \begin{tabular}[c]{@{}c@{}}64x64\\ x128\end{tabular}   & upsampling4    & \begin{tabular}[c]{@{}c@{}}2x2 upsample of conv8\_2\\ concatenate with conv1\_2\end{tabular} & \begin{tabular}[c]{@{}c@{}}512x512\\ x96\end{tabular}  \\ \hline
conv4\_1       & \begin{tabular}[c]{@{}c@{}}3x3x256;\\  LeakyReLU\end{tabular}                & \begin{tabular}[c]{@{}c@{}}64x64\\ x256\end{tabular}   & conv9\_1       & \begin{tabular}[c]{@{}c@{}}3x3x32; dilaton rate 3; \\ LeakyReLU\end{tabular}                 & \begin{tabular}[c]{@{}c@{}}512x512\\ x32\end{tabular}  \\ \hline
conv4\_2       & \begin{tabular}[c]{@{}c@{}}3x3x256;\\  LeakyReLU\end{tabular}                & \begin{tabular}[c]{@{}c@{}}64x64\\ x256\end{tabular}   & conv9\_2       & \begin{tabular}[c]{@{}c@{}}3x3x32; dilaton rate 3; \\ LeakyReLU\end{tabular}                 & \begin{tabular}[c]{@{}c@{}}512x512\\ x32\end{tabular}  \\ \hline
pool4          & 2x2 max pool; stride 2                                                       & \begin{tabular}[c]{@{}c@{}}32x32\\ x256\end{tabular}   & conv10         & 1x1x1                                                                                        & \begin{tabular}[c]{@{}c@{}}512x512\\ x1\end{tabular}   \\ \hline
conv5\_1       & \begin{tabular}[c]{@{}c@{}}3x3x512; \\ LeakyReLU\end{tabular}                & \begin{tabular}[c]{@{}c@{}}32x32\\ x512\end{tabular}   &                &                                                                                              &                                                        \\ \hline
conv5\_2       & \begin{tabular}[c]{@{}c@{}}3x3x512; \\ LeakyReLU\end{tabular}                & \begin{tabular}[c]{@{}c@{}}32x32\\ x512\end{tabular}   &                &                                                                                              &                                                        \\ \hline
\end{tabular}
}
\end{table}

\subsection{Loss Weights}
\label{sec:Loss}
Malignant lesions are usually observed with high SUV values ($>2.5$) in PET scans \cite{Kostakoglu}. We note that most of the SUV values in PET scans are low and only a minority include high SUVs. Hence, we used the SUV value in each pixel as a weight for the pixel-wise loss function as in equation (\ref{eq:weights}), where $N$ is the number of samples. By this we allow the network to pay more attention to high SUVs even though most pixels include lower values. 
\begin{equation}
\label{eq:weights}
L = \frac{1}{N}\sum_{i=1}^{N} I_{PET}(i)(Syn_{PET}(i)-I_{PET}(i))^2
\end{equation}
While this approach helped the FCN to learn the malignant tumor appearance and provide a better response in the synthesized PET images, it did not help when training the cGAN. Hence, we modified the loss function by computing the weighted average reconstruction loss (equation \ref{eq:weights}) for high SUVs ($>2.5$) and for low SUVs ($\leq 2.5$) as in equation (\ref{eq:new_weights}). This way the cGAN training was able to achieve better response in regions with high SUV while not substantially reducing the quality of reconstruction in other regions.
\begin{equation}
\label{eq:new_weights}
L_{cGAN}=L_{low SUV} + L_{high SUV}
\end{equation}

\begin{table}
\centering
\caption{The discriminator used in the proposed cGAN architecture.}
\label{tab:discriminator}
\resizebox{0.65\columnwidth}{!}{%
\begin{tabular}{|c|c|c|}
\hline
\multicolumn{3}{|c|}{Discriminator}                                                                                 \\ \hline
Layer & Details                                                                                        & Size       \\ \hline
input & \begin{tabular}[c]{@{}c@{}}Includes: real/fake PET image;\\ CT image;FCN's output\end{tabular} & 512x512x3  \\ \hline
conv1 & 3x3x32; stride 2; LeakyReLU                                                                    & 256x256x32 \\ \hline
conv2 & 3x3x64; stride 2; LeakyReLU                                                                    & 128x128x64 \\ \hline
conv3 & 3x3x128; stride 2; LeakyReLU                                                                   & 64x64x128  \\ \hline
conv4 & 3x3x256; stride 1; LeakyReLU                                                                   & 64x64x256  \\ \hline
dense & 2 classes (fake/real);softmax                                                                  & 2          \\ \hline
\end{tabular}
}
\end{table}

\section{Experiments and Results}
\label{sec:experiments}
To evaluate the performance of the system, we conducted several sets of experiments. A development set was used for training and validation. Final results are presented for an independent testing set. We demonstrate the applicativity of the system 
to augment performance of a liver lesion detection system. 

\subsection{Dataset}
\label{sec:data}
An institutional review board (IRB) approval was granted for this retrospective study and informed consent was waived by the IRB committee.
The data used in this work includes PET/CT scans (pairs of contrast enhanced portal phase CT scans with their corresponding PET scans) from the Sheba Medical Center, obtained from 2014 to 2015. The dataset contains 60 CT (with 0.97 mm pixel spacing and 4 mm slice thickness) and PET (with 3 mm pixel spacing and 4 mm slice thickness) pairs (from 60 different patients) which we constrained to slices in the region of the liver for our study. Not all PET/CT scans in our dataset included liver lesions. The data was collected in two phases. In the first phase the collected data included PET/CT scans that were used for the development and validation of the explored methods. In the second phase new data was collected for testing with no additional modifications of the algorithms. The training set included 23 PET/CT pairs (6 with malignant liver lesions) and the testing was performed on 37 pairs (9 with malignant liver lesions).  

\subsection{Experimental Setting}
\label{sec:exp}
The networks were implemented and trained using Keras framework \cite{keras} on a PC with a single NVIDIA 1080 Ti GPU. The following hyper-parameters were used for all networks: learning rate of $0.00001$ with a batch size of $4$. Adam optimizer \cite{Kingma} was used with $\beta = 0.5$. For the cGAN, we used $\lambda = 20$ in the optimization process presented in equation \ref{eq:final_optimization}. To assist the network model to be more robust to variability in location and scaling,  online data transformations were performed with uniform sampling of scale [-0.9,1.1] and translations [-25,25] in each epoch. We randomly chose 20\% of the training images for validation which were used to optimize the training process.

\subsection{Reconstruction Evaluation}
\label{sec:reconstruction}
To quantitatively evaluate our method performance in means of reconstruction we used the mean absolute error:
\begin{equation}
\label{eq:MAE}
MAE = \frac{1}{N}\sum_{i=1}^{N} |Syn_{PET}(i)-I_{PET}(i)|
\end{equation}
where $i$ iterates over aligned voxels in the real and synthesized PET images.

In addition, the peak-signal-to-noise-ratio (PSNR) as in \cite{Nie,Wolterink} was used:
\begin{equation}
\label{eq:PSNR}
PSNR = 10\log_{10}\frac{20^2}{MSE}
\end{equation}
where $MSE$ is the mean-squared error, i.e. $\frac{1}{N}\sum_{i=1}^N(Syn_{PET}(i)-I_{PET})^2$.

Our goal is to get very good reconstruction in the lesion areas, while  keeping high reconstruction quality within the entire scan. High SUV values within a PET scan often serve as an indicator for the malignant lesions. We therefore measure the reconstructive error for the  high SUV values (larger than 2.5) and the low SUV values, as two separate sets. The average of these measures is computed as the final score. Table \ref{tab:comparison} shows quantitative comparison across several possible reconstruction schemes. Detailed description of the rows of the table is provided next:

\textbf{FCN.} In the first set of experiments we tested four different fully convolutional networks: 1) The U-net based model as in Table \ref{tab:generator}; 2) The FCN-4s as in Figure \ref{fig:FCN}; 3) The FCN-8s that does not make use of the pool2 layer in the upsampling path as in \cite{Shelhamer}; 4) The FCN-2s that additionally uses the pool1 layer for the upsampling path. Table \ref{tab:comparison} shows that the U-Net results in  larger reconstruction error for high SUV regions compared to the other tested FCNs. The FCN-8s, FCN-4s, and FCN-2s achieved similar results. In previous works the FCN-4s showed promising performance in the liver lesion detection task including small lesions \cite{FCN,Detection}. Hence we select to use the FCN-4s in the proposed solution. We used the loss as in equation \ref{eq:weights} for training. 

\textbf{cGAN.} The second component of the proposed method is the cGAN. We compared two cGAN  architecture variations: 1) Using the U-Net based generator as in Table \ref{tab:generator} (``GAN-U-Net gen."); 2) Using the FCN-4s as a generator ( ``GAN-FCN-4s gen.").  Table \ref{tab:comparison} indicates 
%that show that the GAN-U-Net gen. had a 
better reconstruction performance in the low SUV regions when using the ``GAN-U-Net gen.". We therefore use it as a refinement step to the the FCN  output that had better reconstruction performance in the high SUV regions. In addition, the ``GAN-U-Net gen." had faster convergence compared to the ``GAN-FCN-4s gen.". We used the loss as in equation \ref{eq:new_weights} for training.

\textbf{Combined.} Our proposed method  combines the FCN-4s and the cGAN. We test it next using three different loss functions:   loss presented in equation \ref{eq:new_weights}, loss presented in equation \ref{eq:weights}, and the standard $L2$ loss. In addition, we  compared the results to  pyramid-based image blending \cite{PET}.
%to combine the FCN-4s output with the cGAN \cite{PET}. 
From Table \ref{tab:comparison} we see that the  proposed method showed superiority over the other methods with an average MAE of 0.72 and 0.79, and PSNR of 30.22 and 30.4 using the loss as in equation \ref{eq:new_weights} and equation \ref{eq:weights}, respectively. When using the loss as in equation \ref{eq:weights} instead of equation \ref{eq:new_weights} our method achieved a better average PSNR, however, we preferred to use the latter since it achieved better reconstruction measurements for the high SUVs.

%\rotatebox[origin=c]{90}{combined}

\begin{table*}
\centering

\caption{Average reconstruction performance for low and high SUV regions using different methods. In bold - the highest scores in each column.}
\label{tab:comparison}
\resizebox{\textwidth}{!}{%
\begin{tabular}{c|c|cc|cc|cc|}
\cline{2-8}
                                                & \multirow{2}{*}{Method} & \multicolumn{2}{c|}{High SUV}                        & \multicolumn{2}{c|}{Low SUV}                        & \multicolumn{2}{c|}{Average Score}                  \\ \cline{3-8} 
                                                &                         & \multicolumn{1}{c|}{MAE} & PSNR                      & \multicolumn{1}{c|}{MAE} & PSNR                     & \multicolumn{1}{c|}{MAE} & PSNR                     \\ \hline
\multicolumn{1}{|c|}{\multirow{4}{*}{\rotatebox[origin=c]{90}{Combined}}} & *FCN-4s-cGAN Eq. (\ref{eq:new_weights})     & \textbf{1.33 $\pm$ 0.65} & 22.40 $\pm$ 2.92          & 0.11 $\pm$ 0.04          & 38.04 $\pm$ 1.92         & \textbf{0.72 $\pm$ 0.35} & 30.22 $\pm$ 2.42         \\ \cline{2-2}
\multicolumn{1}{|c|}{}                          & *FCN-4s-cGAN Eq. (\ref{eq:weights})     & 1.48 $\pm$ 0.66          & 21.70 $\pm$ 2.95          & \textbf{0.09 $\pm$ 0.05} & \textbf{39.1 $\pm$ 1.95} & 0.79 $\pm$ 0.36          & \textbf{30.4 $\pm$ 2.45} \\ \cline{2-2}
\multicolumn{1}{|c|}{}                          & FCN-4s-cGAN L2          & 1.55 $\pm$ 0.66          & 21.10 $\pm$ 2.94          & 0.10 $\pm$ 0.04          & 39.03 $\pm$ 1.94         & 0.83 $\pm$ 0.35          & 30.07 $\pm$ 2.44         \\ \cline{2-2}
\multicolumn{1}{|c|}{}                          & Blending                & 1.50 $\pm$ 0.63          & 21.40  $\pm$ 2.94         & 0.10 $\pm$ 0.04          & 39.00 $\pm$ 2.03         & 0.80 $\pm$ 0.34          & 30.20 $\pm$ 2.49         \\ \cline{1-2}
\multicolumn{1}{|c|}{\multirow{2}{*}{{\rotatebox[origin=c]{90}{cGAN}}}}     & cGAN-U-Net gen.         & 1.70 $\pm$ 0.61          & 20.62 $\pm$ 2.92          & 0.10 $\pm$ 0.04          & 39.06 $\pm$ 1.90         & 0.90 $\pm$ 0.33          & 29.84 $\pm$ 2.41         \\ \cline{2-2}
\multicolumn{1}{|c|}{}                          & cGAN-FCN-4s gen.        & 1.52 $\pm$ 0.63          & 21.10 $\pm$ 3.10          & 0.12 $\pm$ 0.04          & 37.60 $\pm$ 1.95         & 0.82 $\pm$ 0.34          & 29.35 $\pm$ 2.53         \\ \cline{1-2}
\multicolumn{1}{|c|}{\multirow{4}{*}{\rotatebox[origin=c]{90}{FCN}}}      & FCN-4s                  & \textbf{1.33 $\pm$ 0.59} & \textbf{22.50 $\pm$ 2.93} & 0.16 $\pm$ 0.05          & 37.60 $\pm$ 1.99         & 0.74 $\pm$ 0.32          & 30.05 $\pm$ 2.46         \\ \cline{2-2}
\multicolumn{1}{|c|}{}                          & FCN-8s                  & \textbf{1.33 $\pm$ 0.57} & 22.45 $\pm$ 2.92          & 0.15 $\pm$ 0.05          & 37.63 $\pm$ 1.99         & 0.74 $\pm$ 0.31          & 30.04 $\pm$ 2.46         \\ \cline{2-2}
\multicolumn{1}{|c|}{}                          & FCN-2s                  & 1.37 $\pm$ 0.62          & 22.42 $\pm$ 3.02          & 0.14 $\pm$ 0.05          & 37.70 $\pm$ 2.02         & 0.76 $\pm$ 0.34          & 30.06 $\pm$ 2.52         \\ \cline{2-2}
\multicolumn{1}{|c|}{}                          & U-Net                   & 1.52 $\pm$ 0.67          & 21.57 $\pm$ 3.1           & 0.12 $\pm$ 0.04          & 38.56 $\pm$ 1.74         & 0.82 $\pm$ 0.36          & 30.07 $\pm$ 2.42         \\ \hline
\end{tabular}
}
\raggedright *Proposed method
\end{table*}

\begin{figure}
\centering
\includegraphics[width=0.5\textwidth]{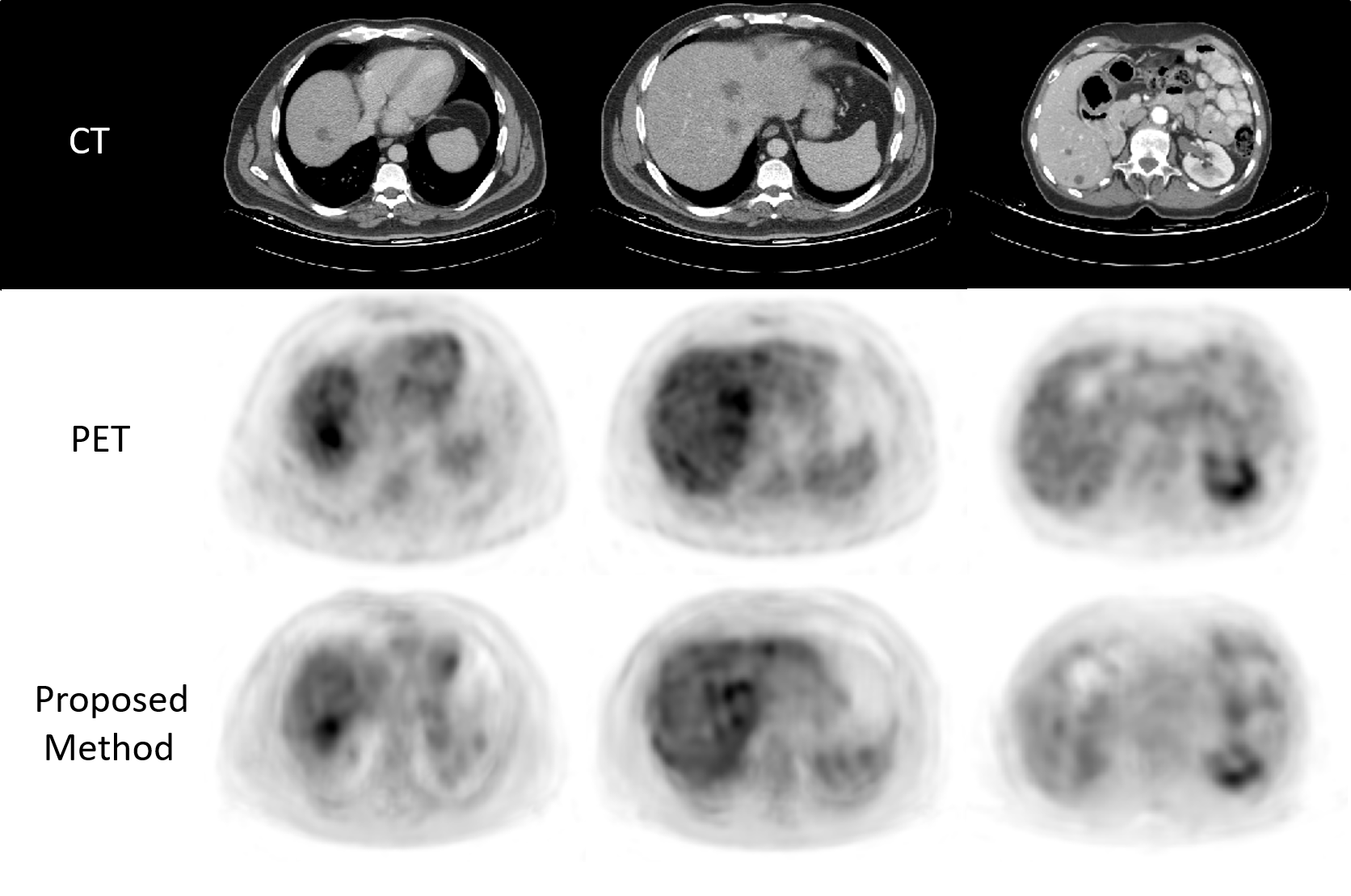}
\caption{Sample results of the predicted PET using our method compared to the real PET with the corresponding CT images.}
\label{fig:QualitativeResults}
\end{figure}

\begin{figure}
\centering
\includegraphics[width=0.5\textwidth]{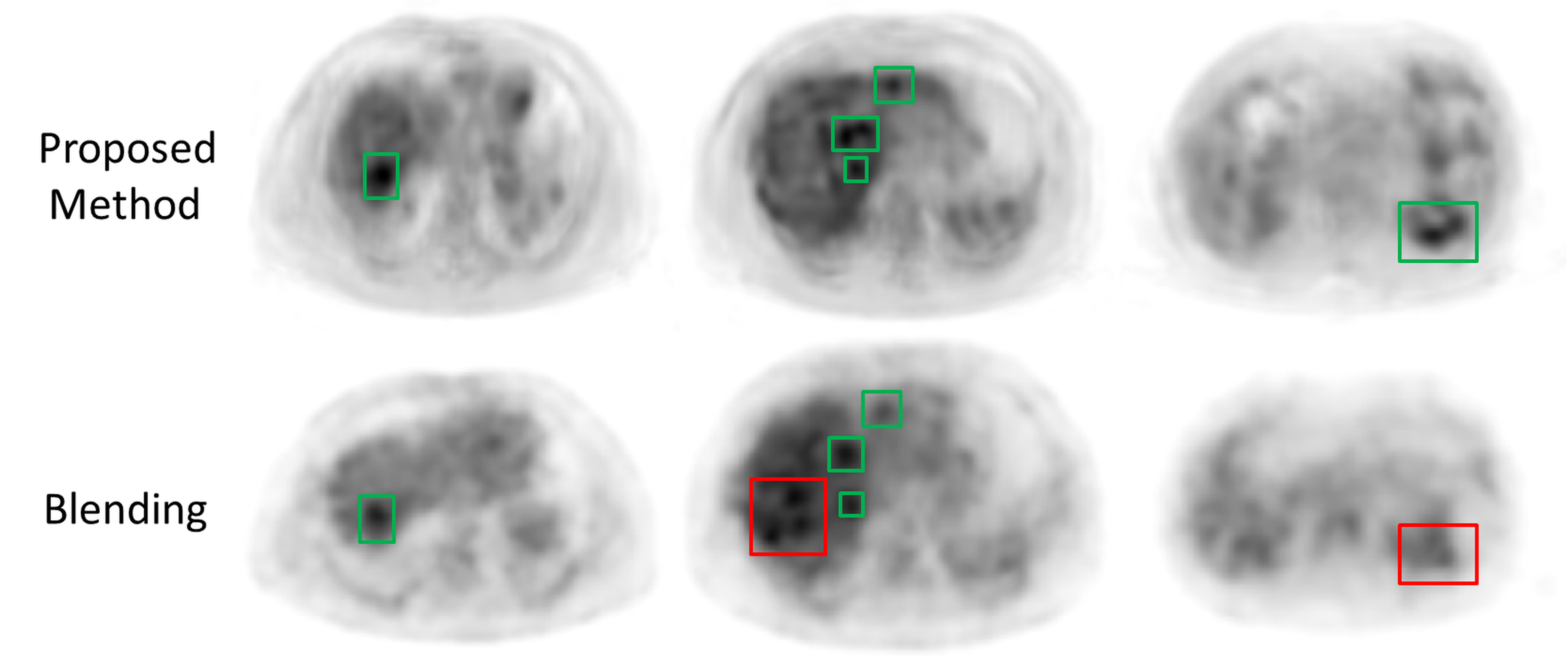}
\caption{Sample results of the predicted PET using our method compared to the image blending based method \cite{PET}. In green - correctly synthesized high SUV regions; In red - false synthesized high SUV regions.}
\label{fig:CompareBlending}
\end{figure}

Qualitative results are shown in Figure \ref{fig:QualitativeResults}. It compares the method's virtual PET images with the original PET study images. The virtual PET provided a very similar response to the real PET in the presented cases. The left column includes one malignant lesion, the second column includes three liver metastases (malignant lesions), and the right column includes two cysts (benign lesions).
In Figure \ref{fig:CompareBlending} we use the same sample cases as in Figure \ref{fig:QualitativeResults} and have compared our proposed method to the image blending based method \cite{PET}. In the left column we can see that the malignant lesion has been recognized using both methods as a dark blob which seems slightly darker and larger using our method. In the second column the three malignant lesions were recognized using both methods, however, the image blending based method included several blobs that are false-positives marked in red. In the right column there are two cysts (benign lesions), and both methods did not have high response as expected, however, our proposed method seems to have a better result for the surrounding tissues such as the left kidney marked in green. 

\subsection{ Liver Lesion Detection using the Virtual-PET}
\label{sec:detection}
In the following experiment, we use the synthesized (Virtual) PET images as an additional false-positive reduction layer for an existing lesion detection software. Since high SUV values can be indicative of malignant lesions, thresholding the PET using a high SUV threshold ($th = 2.5$) can reduce non-relevant regions for lesion detection.
 We use a particular lesion detection system, that was developed in our group \cite{Detection}, which combines global context via an FCN, along with local patch level analysis using superpixel sparse based classification. This framework is made up of two main modules. The first module is an FCN having three  slices are as input: the target slice in the center and two adjacent slices above and below. Using an FCN - based analysis, this module outputs a lesion probability map. Based on the  high-probability candidate lesion regions from the first module, the second module follows with localized  patch level analysis using superpixel sparse based classification. This module's objective is to classify each localized superpixel as a lesion or not. Thus it provides a fine-tuning step, with the objective of increasing sensitivity to lesions while removing false positives (FPs). 

\begin{figure}
\centering
\includegraphics[width=0.5\textwidth]{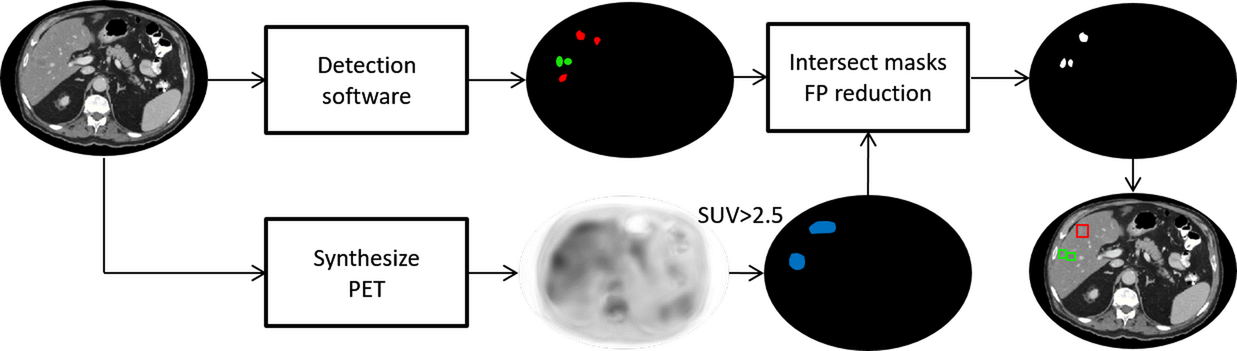}
\caption{Combining our proposed method to synthesize PET with a lesion detection software: 
The output of the detection software includes true positives (in green) along with false positives (in red). Thresholding over the synthesized PET image to extract high-response regions (in blue) can help reduce the false positives by intersecting the detection mask with the thresholding mask.}
\label{fig:detectionFramework}
\end{figure}

We suggest the following scheme for improving the lesion detection software: Given a CT scan, the detection software outputs lesion candidates as a binary mask which may include false detections. By thresholding the synthesized PET scan and finding the intersection of both the detection software's candidates mask and the PET thresholding result some of the false detections can be removed. Figure \ref{fig:detectionFramework} illustrates this process. This is a rather na\"ive approach that shows the clinical relevance of using the virtual PET to improve existing software.

An additional test set of 14 CT scans including 55 lesions was used for the following experiments. Two evaluation measurements were computed, the true positive rate (TPR) and false positive rate (FPR) for each case as follows:
\begin{itemize}
\item $TPR$- Number of correctly detected lesions divided by the total number of lesions.
\item $FPR$- Number of false positives per scan.
\end{itemize}

Table \ref{tab:detection} shows the performance with and without the proposed false-positive reduction layer (note that the results were constrained to the manually annotated liver). Using our method the average FPR decreased from 2.9 to 2.1 (improvement of 28\% with P-value$<$0.05) with a similar TPR.

\begin{table}
\centering
\caption{Detection measurements with and without SUV thresholding on the synthesized PET. In bold - the results obtained using the proposed method.}
\label{tab:detection}
\begin{tabular}{|c|c|c|}
\hline
\textbf{Method}                          & \textbf{TPR{[}\%{]} } & \textbf{Average FPR}\\ \hline
Detection soft.                 & 94.6 & 2.9 $\pm$ 2.1         \\ \hline
Detection soft+ proposed & \textbf{94.6} & \textbf{2.1 $\pm$ 1.7}         \\ \hline
Detection soft+ blending & 90.9 & 2.2  $\pm$ 1.7      \\ \hline
Detection soft+ FCN-4s & 90.9 & 2.2 $\pm$ 1.7        \\ \hline

\end{tabular}
\end{table}

\begin{figure}
\centering
\includegraphics[width=0.5\textwidth]{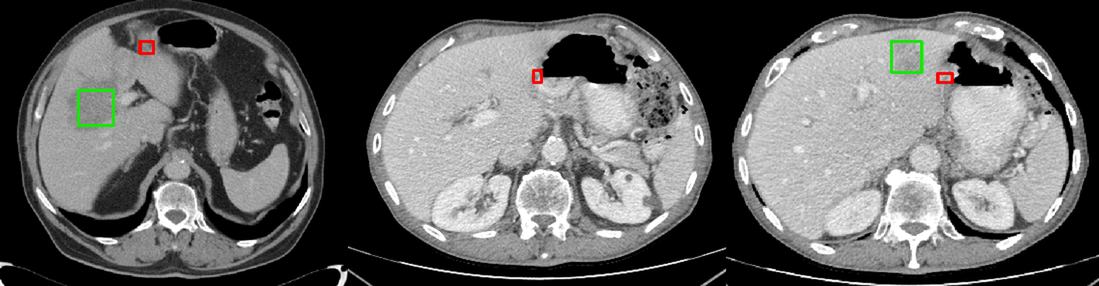}
\caption{Sample results using the existing detection software. In green - correctly detected lesions. In red - false positives that were removed by combining the proposed method.}
\label{fig:detectionExamples}
\end{figure}

Figure \ref{fig:detectionExamples} shows examples of cases with false positives that were removed (in red) by combining the proposed method with the existing detection software.

\begin{figure}
\centering
\includegraphics[width=0.5\textwidth]{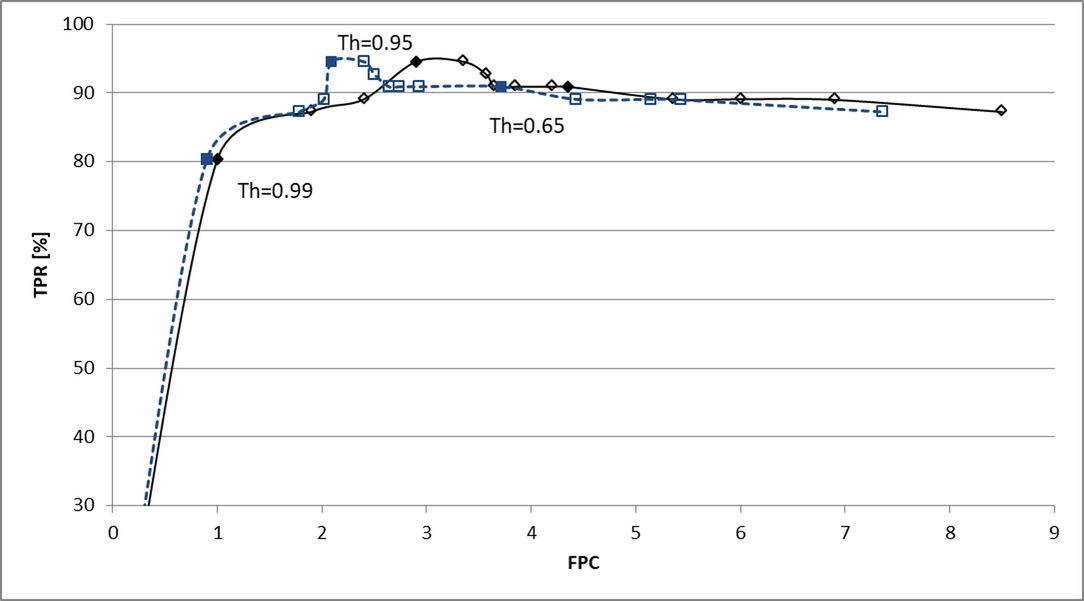}
\caption{FROC curve of lesion detection using FCN with sparsity based FP reduction (solid black line) and with the addition of synthesized PET FP reduction (dashed blue line).}
\label{fig:FROC}
\end{figure}

One key parameter of the detection system is the probability threshold, \textit{th}, which defines the set of candidate regions extracted from the FCN probability maps. We compare the free-response receiver operating characteristic (FROC) presented in \cite{Detection} with the FROC achieved using the combination of the synthesized PET with the current system (Figure \ref{fig:FROC}). It can be seen that our system was able to reduce the amount of false positives in each tested \textit{th} while preserving the TPR. Unlike the classic FROC curve where the TPR increases for decreasing \textit{th}, here there was a moderate decrease for \textit{th} below 0.95. The TPR can decrease when two candidates merge into one candidate even though there are two lesions. Note that We used $th = 0.95$ in our experiments (table \ref{tab:detection} and figure \ref{fig:detectionExamples}) as in the original system. 

As a final experiment, we tested 
a fully automatic framework with an automatic liver segmentation scheme, as in \cite{Detection}.  Thus, we use an automatic liver segmentation instead of the manually circumscribed liver. Using our method the average FPR decreased from 3.0 to 2.3 (improvement of 23\% with P-value$<$0.05) with a slight decrease of the TPR from 90.9\% to 89.1\% (not significant).

\section{Discussion}
\label{sec:discussion}
A novel system for PET synthesis using only CT scans has been presented. The presented system includes an FCN model and a cGAN model that refines the synthesized output extracted from the FCN. This framework provides a realistic PET estimation with special attention to malignant lesions using a custom loss function for each model.

Table \ref{tab:comparison} shows the reconstruction performance in terms of MAE and PSNR for high SUV regions, low SUV regions, and an average of both which is an estimate of the balance we want to have between these regions. The high SUV regions are important since they usually signify malignant lesions inside the liver, but  it is important to reconstruct the low SUV regions as well, since we want to have a good contrast between malignant and non-malignant tissues. The FCN-4s achieved the best PSNR and MAE for the high SUV regions with our proposed method achieving very close performance measures. However, for the low SUV regions, the FCN-4s got inferior PSNR and MAE, while the cGAN and the image blending got the best results. Using the average scores our proposed method seems to have the best balance between the high and low SUV regions.

In an additional experiment we wanted to see if the proposed method can be used to improve an existing automatic liver lesion detection software \cite{Detection}. Our system was easily integrated into the lesion detection software. Using a pathological SUV threshold of 2.5 we achieved a decrease in false-positive from an average of 2.9 per case to 2.1 (28\% improvement). This experiment shows the benefit of using the proposed system to improve a given liver lesion analysis software. Since this method was trained on a dataset that was not seen by the existing detection software it improved its results. However, no manual labeling was conducted in this experiment since our method uses only PET/CT pairs for training.

The current work includes several limitations, which we discuss next. The experiments conducted in this study focused on the liver region. The liver is a common site for metastases in oncological patients, hence,  more examples of malignancies can be extracted and can help the training process. More work should be done to make use of the proposed framework for other malignancies within different regions in the CT scan, preferably with no manually annotated labels. Adaptation of this method to other types of malignant tumors may require to re-train it per-organ since different malignant tumors have different appearances in CT images (e.g. lung cancer vs. liver cancer). 
We used online data augmentation including scaling and translation, and random noise addition in the training process. Additional data augmentation techniques such as random distortions and deformations or image rotations can be used to further improve the system performance and robustness but were not found necessary in our experiments. An additional point that could be explored is the combination of the FCN output with the training of the cGAN. We have proposed to concatenate the FCN output to the CT image as input to the cGAN and compared it to a pyramid based image blending approach. However, different merging approaches should be explored within the cGAN generator. For example, using the FCN output as an input to separate group of layers in the network and merge with the CT image deeper in the network model. One of the parameters that we use in our system is the SUV cut-off value. Although our system shows adequate results using 2.5 as the cut-off value, tweaking it may perhaps further optimize our method.

One possible application could be to use the virtual PET to improve lesion segmentation. However, the PET images are quite blurry and so is the virtual PET, making it hard  to assess the segmentation process. Hence, we believe that detection approaches are more relevant for this method. 

We used a rather na\"ive approach by thresholding the virtual PET to reduce the amount of false-positives per case in an existing lesion detection software. However, the proposed system can be easily integrated into the training process of different networks for different tasks such as detection (as shown here) and classification.

To conclude, our proposed framework with  the FCN-cGAN combination and the custom loss function has shown promising results in terms of reconstruction measures as well as detection measures by integrating it with an existing lesion detection software.
A major strength of this paper is that no manual labeling was used to train the system. As we well know, the task of manually labeling and annotating medical data is hard, and contributes to the usually small data sets that are used in todays’ medical imaging research. Each year millions of PET/CT studies are conducted worldwide, and utilizing the current method, the CT and PET pairing can be used as “free” labeled and annotated data, with potential for big data, approaching millions of studies.
Future work entails obtaining a larger dataset with vast experiments using the entire CT and not just the liver region as well as integrating it into the training process of deep learning based detection and classification networks. The presented system can be used for many applications in which PET examination is needed such as evaluation of drug therapies and detection of malignant lesions.

\subsubsection*{Acknowledgment}
This research was supported by the Israel Science Foundation (grant No. 1918/16).

Avi Ben-Cohen's scholarship was funded by the Buchmann Scholarships Fund.

\subsection*{Disclosure of conflict of interest}
The authors have no relevant conflicts of interest to disclose.

% \section*{References}

\bibliographystyle{elsarticle-num}
% \bibliography{mybib}

\begin{thebibliography}{4}
\bibitem{Adelson} Adelson, E. H., Anderson, C. H., Bergen, J. R., Burt, P. J., and Ogden, J. M.,1984. Pyramid methods in image processing. RCA engineer, 29(6), pp. 33-41.

\bibitem{FCN} Ben-Cohen, A., Diamant, I., Klang, E., Amitai, M., and Greenspan, H., 2016. Fully Convolutional Network for Liver Segmentation and Lesions Detection. In International Workshop on Large-Scale Annotation of Biomedical Data and Expert Label Synthesis. Springer International Publishing, pp. 77-85.

\bibitem{Detection} Ben-Cohen, A., Klang, E., Kerpel, A., Konen, E., Amitai, M. M., and Greenspan, H., 2017. Fully convolutional network and sparsity-based dictionary learning for liver lesion detection in CT examinations. Neurocomputing.

\bibitem{PET} Ben-Cohen, A., Klang, E., Raskin, S. P., Amitai, M. M., and Greenspan, H.,2017. Virtual PET Images from CT Data Using Deep Convolutional Networks: Initial Results. In International Workshop on Simulation and Synthesis in Medical Imaging. Springer, Cham, pp. 49-57.

\bibitem{Bi} Bi, L., Kim, J., Kumar, A., Feng, D., and Fulham, M., 2017. Synthesis of Positron Emission Tomography (PET) Images via Multi-channel Generative Adversarial Networks (GANs). In Molecular Imaging, Reconstruction and Analysis of Moving Body Organs, and Stroke Imaging and Treatment. Springer, Cham, pp. 43-51.

\bibitem{Chartsias} Chartsias, A., Joyce, T., Dharmakumar, R., and Tsaftaris, S. A., 2017. Adversarial Image Synthesis for Unpaired Multi-modal Cardiac Data. In International Workshop on Simulation and Synthesis in Medical Imaging. Springer, Cham, pp. 3-13.

\bibitem{keras} Chollet, Fran\c{c}ois et al.: Keras. \url{https://github.com/keras-team/keras}. GitHub, (2015).

\bibitem{Christ} Christ, P. F., Ettlinger, F., Grün, F., Elshaera, M. E. A., Lipkova, J., Schlecht, S., ... and Rempfler, M., 2017. Automatic Liver and Tumor Segmentation of CT and MRI Volumes using Cascaded Fully Convolutional Neural Networks. arXiv preprint arXiv:1702.05970.

\bibitem{Goodfellow} Goodfellow, I., Pouget-Abadie, J., Mirza, M., Xu, B., Warde-Farley, D., Ozair, S., and Bengio, Y.,2014. Generative adversarial nets. In Advances in neural information processing systems, pp. 2672-2680.

\bibitem{Han} Han, X.,2017. MR based synthetic CT generation using a deep convolutional neural network method. Medical Physics, 44(4), pp. 1408-1419.

\bibitem{Higashi} Higashi, K., Clavo, A. C., and Wahl, R. L.,1993. Does FDG Uptake Measure the Proliferative Activity of Human Cancer Cells? In Vitro Comparison with DNA Flow Cytometry and Tritiated Thymidine Uptake. Journal of Nuclear Medicine, 34, 414-414.

\bibitem{Isola} Isola, P., Zhu, J. Y., Zhou, T., and Efros, A. A.,2016. Image-to-image translation with conditional adversarial networks. arXiv preprint arXiv:1611.07004.

\bibitem{Kelloff} Kelloff, G. J., Hoffman, J. M., Johnson, B., Scher, H. I., Siegel, B. A., Cheng, E. Y., and Shankar, L., 2005. Progress and promise of FDG-PET imaging for cancer patient management and oncologic drug development. Clinical Cancer Research, 11(8), 2785-2808.

\bibitem{Kinehan} Kinehan, P. E., and Fletcher, J. W., 2010. PET/CT standardized uptake values (SUVs) in clinical practice and assessing response to therapy. Semin Ultrasound CT MR, 31(6), 496-505.
\bibitem{Kingma} Kingma, D. P., and Ba, J., 2014. Adam: A method for stochastic optimization. arXiv preprint arXiv:1412.6980.

\bibitem{Kostakoglu} Kostakoglu, L., Agress Jr, H., and Goldsmith, S. J., 2003. Clinical role of FDG PET in evaluation of cancer patients. Radiographics, 23(2), 315-340.

\bibitem{LeCun}	LeCun, Y., Bottou, L., Bengio, Y., and Haffner, P.,1998. Gradient-based learning applied to document recognition. Proceedings of the IEEE, 86(11), 2278-2324.

\bibitem{Mets} Metz, C. E., 2006. Receiver operating characteristic analysis: a tool for the quantitative evaluation of observer performance and imaging systems. Journal of the American College of Radiology, 3(6), 413-422.

\bibitem{Nie} Nie, D., Cao, X., Gao, Y., Wang, L., and Shen, D., 2016. Estimating CT image from MRI data using 3D fully convolutional networks. In International Workshop on Large-Scale Annotation of Biomedical Data and Expert Label Synthesis. Springer International Publishing, pp. 170-178.

\bibitem{Ronneberger} Ronneberger, O., Fischer, P., and Brox, T.,2015. U-net: Convolutional networks for biomedical image segmentation. In International Conference on Medical Image Computing and Computer-Assisted Intervention, Springer International Publishing, pp. 234-241.

\bibitem{Shelhamer} Shelhamer, E., Long, J., and Darrell, T., 2016. Fully convolutional networks for semantic segmentation. IEEE transactions on pattern analysis and machine intelligence.

\bibitem{Simonyan} Simonyan, K., and Zisserman, A., 2014. Very deep convolutional networks for large-scale image recognition. arXiv preprint arXiv:1409.1556.

\bibitem{Weber} Weber, W. A., Grosu, A. L., and Czernin, J., 2008. Technology Insight: advances in molecular imaging and an appraisal of PET/CT scanning. Nature Clinical Practice Oncology, 5(3), 160-170.

\bibitem{Wolterink} Wolterink, J. M., Dinkla, A. M., Savenije, M. H., Seevinck, P. R., van den Berg, C. A., and Išgum, I., 2017. Deep MR to CT synthesis using unpaired data. In International Workshop on Simulation and Synthesis in Medical Imaging. Springer, Cham, pp. 14-23.

\bibitem{Weber2} Weber, W. A., 2009. Assessing tumor response to therapy. Journal of nuclear medicine, 50(Suppl 1), 1S-10S.

\bibitem{Xiang} Xiang, L., Wang, Q., Nie, D., Qiao, Y., and Shen, D., 2017. Deep Embedding Convolutional Neural Network for Synthesizing CT Image from T1-Weighted MR Image. arXiv preprint arXiv:1709.02073.

\end{thebibliography}

\end{document}